\pdfoutput=1

\documentclass{article}

\usepackage[preprint]{neurips_2023}

\usepackage[utf8]{inputenc} 
\usepackage[T1]{fontenc}    
\usepackage{hyperref}       
\usepackage{url}            
\usepackage{booktabs}       
\usepackage{amsfonts}       
\usepackage{nicefrac}       
\usepackage{microtype}      
\usepackage{xcolor}         
\usepackage{multirow,multicol}
\usepackage{graphicx}
\usepackage{enumitem}
\usepackage{amsmath}
\usepackage{subcaption}
\usepackage{float}
\usepackage{listings}
\usepackage{tcolorbox}
\usepackage{tikz}
\usepackage{geometry}
\usepackage{rotating}

\title{AI-native Memory:\\A Pathway from LLMs Towards AGI}

\author{
  Jingbo Shang$\quad$ Zai Zheng$^*\quad$ Jiale Wei$^*\quad$ Xiang Ying$\quad$ Felix Tao$\quad$ Mindverse Team\\
  Mindverse AI \\
  \{yingxiang, tao\}@mindverse.ai
}

\usepackage{xspace}

\newcommand\blfootnote[1]{
    \begingroup
    \renewcommand\thefootnote{}\footnote{#1}
    \addtocounter{footnote}{-1}
    \endgroup
}

\newcommand{\smallsection}[1]{\noindent\textbf{#1}.}

\begin{document}

\maketitle

\begin{abstract}
    Large language models (LLMs) have demonstrated the world with the sparks of artificial general intelligence (AGI). 
One opinion, especially from some startups working on LLMs, argues that an LLM with nearly unlimited context length can realize AGI. 
However, they might be too optimistic about the long-context capability of (existing) LLMs --
(1) Recent literature has shown that their effective context length is significantly smaller than their claimed context length;
and (2) Our reasoning-in-a-haystack experiments further demonstrate that simultaneously finding the relevant information from a long context and conducting (simple) reasoning is nearly impossible. 
In this paper, we envision a pathway from LLMs to AGI through the integration of \emph{memory}. 
We believe that AGI should be a system where LLMs serve as core processors. 
In addition to raw data, the memory in this system would store a large number of important conclusions derived from reasoning processes.
Compared with retrieval-augmented generation (RAG) that merely processing raw data, this approach not only connects semantically related information closer, but also simplifies complex inferences at the time of querying. 
As an intermediate stage, the memory will likely be in the form of natural language descriptions, which can be directly consumed by users too.
Ultimately, every agent/person should have its own lifelong personal model, a deep neural network model (thus \emph{AI-native}) that parameterizes and compresses all types of memory, even the ones cannot be described by natural languages.
Finally, we discuss the significant potential of AI-native memory as the transformative infrastructure for (proactive) engagement, personalization, distribution, and social in the AGI era, as well as the incurred privacy and security challenges with preliminary solutions.

    \blfootnote{* Equal Contribution}
\end{abstract}

\section{Introduction}

Large language models (LLMs), pre-trained on massive text corpora and instruction-tuned on expert annotations (and also via reinforcement learning with human feedback), such as
the GPT series from OpenAI~\citep{brown2020language,ouyang2022training,achiam2023gpt}, 
the Gemini series from Google~\citep{team2023gemini,reid2024gemini,team2024gemma},
the Claude series from Anthropic~\citep{anthropic2024claude},
the Llama series from Meta~\citep{touvron2023llama,touvron2023llama2}, and
the Mixtral series from Mistral~\citep{jiang2023mistral,jiang2024mixtral},
have demonstrated significant potentials in their capabilities as general task solvers, going beyond language modeling itself. 
These models can follow complicated human instructions and perform multi-step reasoning when necessary~\citep{wei2022chain,zeng2023evaluating}. 
Therefore, it is a consensus that LLMs are becoming fundamental building blocks towards artificial general intelligence (AGI)~\citep{bubeck2023sparks,minaee2024large}.

Long-context processing capability is vital for LLMs, and therefore, is one of the most popular directions in LLM researches. 
For example, the original GPT-4 has a context window of 32K tokens~\citep{achiam2023gpt}, and the the most recent GPT-4-turbo and GPT-4o models can process 128K tokens; Gemini 1.5 claimed a context window of 1M or 10M tokens~\citep{reid2024gemini}. 
Academia people have also explored to combat length extrapolation~\citep{peng2023yarn,xiao2023efficient,han2023lm,zhang2024soaring} and position bias~\citep{liu2024lost,peysakhovich2023attention,an2024make}, where some works claimed ``unlimited'' context lengths. 
Following this trend, an increasing number of people, especially from startups working on LLMs, argue that an LLM with super long or even unlimited context can realize AGI by putting all raw data into the context and relying entirely on the LLM to complete all necessary reasoning in one step to get the final result for each query.

While nowadays LLMs can take super or even infinitely long inputs and produce an output without throwing a runtime error, it is still unknown whether these models can appropriately utilize the provided long contexts. 
We argue that similar to a human's cognitive load~\citep{sweller1988cognitive}, the maximum amount of content LLMs are capable of handling might be inherently limited depending on the task they are performing. 
However, most previous evaluations for long-context models are based on perplexity or a simple synthetic retrieval task while overlooking the effectiveness on more complex tasks. 
According to a recent benchmark following more complicated tasks~\citep{hsieh2024ruler}, most, if not all, LLMs over-claimed their context lengths. 
For example, GPT-4 models, which claim to have a context of 128K, only has an \emph{effective context} of 64K; 
ChatGLM~\citep{zeng2023glm-130b,du2022glm}, another model claimed to have a context of 128K, ends up with only 4K.
We further develop reasoning-in-a-haystack evaluations following the LLM-as-personal-assistant scenarios and demonstrate that simultaneously finding the relevant information from a long context and conducting reasoning is nearly impossible. 

We believe that AGI should be a system, where LLMs are more like Processors and LLM's context is like a RAM.
Using Processor and RAM alone is not even enough for a computer, nor AGI. 
To complete this system, we will at least need (long-term) \emph{Memory}, which plays a role of disk storage. 
Retrieval-augmented LLMs that sift through numerous relevant contexts to answer a query~\citep{kovcisky2018narrativeqa,dasigi2021dataset,pang2022quality,trivedi2022musique} can be viewed as a special case here by defining the Memory as raw data only. 
However, Memory is beyond the raw data, as it should be generated and organized, including many results that require reasoning from the raw data. 
In addition to downstream applications, Memory shall be able to be directly consumed by users. 

Acknowledging the necessity of Memory, we then discuss the forms of Memory and how to facilitate the interaction between Memory and LLM (e.g., loading the right data from ``disk'' to ``RAM'').
As an intermediate stage, the memory will likely be in the form of natural language descriptions.
This is in line with many existing information extraction and knowledge discovery works and we will construct a ``Memory Palace'' for each agent/person.
Ultimately, every agent/person should have its own lifelong personal model (LPM), a deep neural network model (thus \emph{AI-native}) that parameterizes and compresses all types of memory, even the ones cannot be described by natural languages.
From this compression perspective, this LPM can be a LLM too. 
Finally, we discuss the significant potential of AI-native memory as the transformative infrastructure for AI-native (proactive) engagement, personalization, distribution, and social in the AGI era, as well as the incurred privacy and security challenges with preliminary solutions.

In summary, our main points are
\begin{itemize}[nosep,leftmargin=*]
    \item LLM itself is not enough for AGI. It is very challenging and even impossible to build an LLM with truly unlimited context length, so the model can put all raw data into the context and complete all necessary reasoning in one step for a particular query.
    \item Memory is a keystone towards AGI. AGI should be a system, where LLMs are more like Processors, LLM's context is like a RAM, and Memory plays a role like a disk. 
    \item There can be at least two different ways to generate and organize Memory. The first solution is following the Information Extraction/Generation ideas of constructing a ``Memory Palace''. The second solution falls in the line of compressing the Memory as a neural network (maybe LLM too).
\end{itemize}
\section{LLMs with Unlimited Context Length are NOT the Answer for AGI}

As the LLMs have demonstrated the world with the sparks of AGI~\citep{bubeck2023sparks}, an increasing number of people, especially from some startups working on LLMs, argue that an LLM with super long or even unlimited context can achieve AGI by putting all raw data into the context and relying entirely on the LLM to complete all necessary reasoning in one step to get the final result.
There are two key assumptions behind this long-context direction, and they \emph{must hold true at the same time}; otherwise, this argument would fail automatically. 
\begin{enumerate}[nosep,leftmargin=25mm]
    \item[Assumption 1:] LLMs can effectively find the necessary information from a super long or even unlimited context, i.e., the \emph{needle-in-a-haystack} capability.
    \item[Assumption 2:] LLMs can conduct all the required, complicated inferences based on the raw inputs in one step, i.e., the \emph{long-context reasoning} capability. 
\end{enumerate}
According to the current literature and our experiments (will be presented in this section), people might be too optimistic about the long-context capability of (existing) LLMs --
(1) recent literature~\citep{hsieh2024ruler} has shown that their effective context length is significantly smaller than their claimed context length;
and (2) our reasoning-in-haystack experiments in Section~\ref{sec:reasoning-in-a-haystack} further demonstrate that simultaneously finding the relevant information from a long context and conducting reasoning is nearly impossible. 
More details will be covered in the remainder of this section.

\subsection{Effective Context Length of Existing LLMs is Limited}

There are several proprietary LLMs claimed very long context lengths.
For example, the original GPT-4 has a context window of 32K tokens~\citep{achiam2023gpt}, and the the most recent GPT-4-turbo and GPT-4o models can process 128K tokens; Gemini 1.5 claimed a context window of 1M or 10M tokens~\citep{reid2024gemini}. 
There are also a number of works, mostly from academia, extending the open-source LLMs to long context lengths, by either adding more fine-tuning with long contexts or modifying the (relative) attention calculations without changing the model parameters~\citep{peng2023yarn,xiao2023efficient,han2023lm,zhang2024soaring,liu2024lost,peysakhovich2023attention,an2024make} 

\smallsection{Needle-in-a-haystack (NIAH)}
The needle-in-a-haystack test is commonly adopted in these long-context LLM works to demonstrate that the LLMs can retrieve the ``needle'' (e.g., a specific number or sentence) from the ``haystack'', i.e., a long irrelevant/background text.

\smallsection{Effective Context Length}
The effective context length is defined as the maximum length that the testing LLM can outperform a strong baseline.
Specifically in ~\citep{hsieh2024ruler}, the baseline is chosen as LLAMA-2-7B (chat), a popular open-source LLM with a 4K context length that is very affordable for serving. 
All the testing LLMs have a claimed context length at least 32K.

According to the Table 3 in~\citep{hsieh2024ruler}, most, if not all, LLMs overclaimed their context lengths. 
For example, GPT-4~\citep{achiam2023gpt}, which claims to have a context of 128K, only has an effective context of 64K; 
ChatGLM~\citep{zeng2023glm-130b,du2022glm} claims to have a context of 128K, but its effective context is only 4K.

Therefore, we believe that super long/unlimited effective context is very difficult to achieve, and the effective context size in existing long-context solutions has not fundamentally improved. 
There are still many fundamental obstacles in technology in the future.

\subsection{Reasoning-in-a-haystack is Very Difficult for Existing LLMs}
\label{sec:reasoning-in-a-haystack}

\begin{figure}[H]
    \centering
    \includegraphics[width=\linewidth]{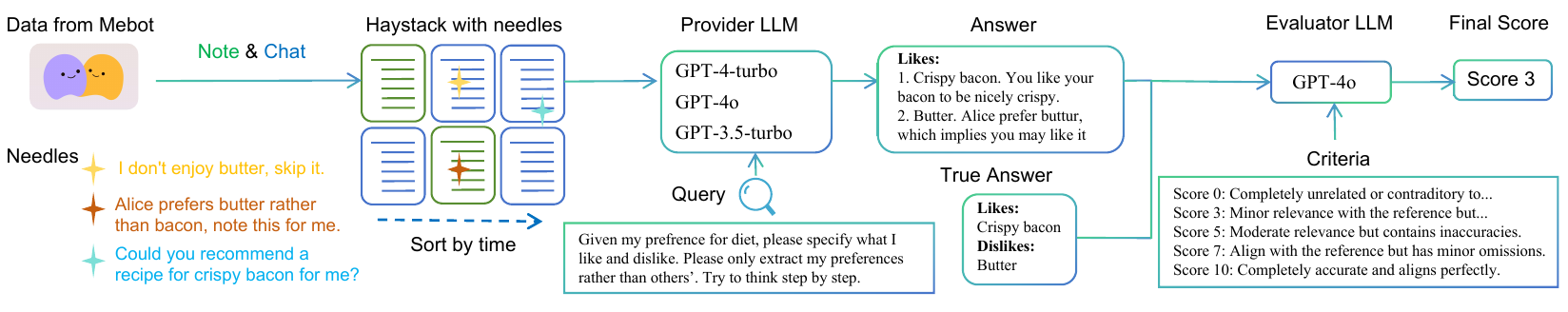}
    \vspace{-3mm}
    \caption{An Overview of Reasoning-in-a-Haystack. In this paper, the haystack, needles, and queries are all designed based on the real data and scenarios from Mebot of Mindverse AI, under the user permission. The haystack is typically a series of User-Mebot interactions chained chronologically. The needle-query pairs are constructed for certain recommendation scenarios.}
    \vspace{-3mm}
    \label{fig:exp_design}
\end{figure}

Going beyond the traditional NIAH tasks that focus solely on retrieval-based abilities, we propose a new reasoning-in-a-haystack task, aimed at validating LLMs' capability when the retrieval and reasoning are required simultaneously.
Figure~\ref{fig:exp_design} shows an overview of the reasoning-in-a-haystack evaluation pipeline.
We start with the real data from Mebot\footnote{\url{https://me.bot/}. We would like to acknowledge to the users who have agreed to our experiment use for their data.} of Mindverse AI.
\textbf{Mebot} is a ``second me'' product based on LLMs. 
For each user, it creates personalized models that can be applied across various scenarios. 
Specifically, it emphasizes on organizing the user's memories while ensuring privacy and security, providing personalized services and inspiration based on these memories.

\subsubsection{Experiment Setups}
The experiment details are described as follows.

\smallsection{Haystack, Needle, and Query: A more challenging setting}
We constructed 8 haystacks for different users to increase the diversity and difficulty of the test cases. 
Each haystack, served as a chronologically organized compilation of users' notes and session messages, was created with the explicit consent of the users. 
The data was sourced from Mebot users and meticulously filtered to ensure the absence of contradictory information in each query-needle pair. 
These data contains \emph{note} and \emph{chat}.
Each note includes title, summary, and content;
each chat session involves a (multi-turn) dialogue between user and Mebot. 

We designed 6 distinct, well-structured query-needle pairs, each with a corresponding true answer, as exemplified in Appendix~\ref{sec:example-multi-needle}. 
All pairs are in the context of Mebot and are close-ended to ensure feasibility for automated evaluation. 
The number of hops, which represents the reasoning steps required to obtain the final result, is set to 1, 2, and 3. 
Furthermore, we experimented two different ways to distribute the needles in the haystack as follows.
\begin{itemize}[nosep,leftmargin=*]
    \item \textbf{Multi-needle}: Every needle is evenly distributed in the haystack. 
    For example, if there are 5 needles, they are placed at depths of 0\%, 20\%, 40\%, 60\%, and 80\%.
    \item \textbf{Single-needle}: All the needles are combined together and distributed at the depth of 40\% or 60\%.
\end{itemize}

Note that our constructed haystack, needle, and query shall be viewed as significantly more challenging than previous NIAH works, where the relevance between haystack and needle-query pair is nearly minimal.

\smallsection{Compared Provider LLMs}
We selected \texttt{GPT-4o}, \texttt{GPT-4-turbo} and \texttt{GPT-3.5-turbo} as the Provider to be evaluated, as \texttt{GPT-4o}, \texttt{GPT-4-turbo} are two of the most advanced models and \texttt{GPT-3.5-turbo} serve as a preferable baseline. 
The prompt settings used for these providers are illustrated in Appendix~\ref{sec:prompt_template}.

\smallsection{Evaluator LLM, True Answer, and Evaluation Criteria}
Due to the closed-world nature of our needle-query construction, 
we first generate a true answer by LLM and then refine it manually to ensure accuracy and fairness for evaluation. 
The introduction of true answer makes the evaluator's job much easier as it only needs to compare provider's answer with well-designed true answer; there is no need to refer to the needles to handle more complex reasoning during the evaluation.
To ensure consistency in our evaluation, we used \texttt{GPT-4o} (temperature=0) as the evaluator for all cases. 
The evaluation criteria are presented in Appendix~\ref{sec:eval_criteria}.
For the same provider LLM, we iterate through all needle-query pairs and conduct experiments on 8 haystacks to obtain an average score, which is a number between 0 and 10, the higher, the better.

\subsubsection{Results}
\begin{figure}[t]
    \centering
    \begin{tikzpicture}
        \node[rotate=90] at (-7, 0.29) {\scriptsize Context length};

        \node at (0, 0) {
            \begin{minipage}{\textwidth}
                \centering
                \subfloat[Multi-Needle (Uniform Depth)]{
                    \includegraphics[width=0.48\linewidth]{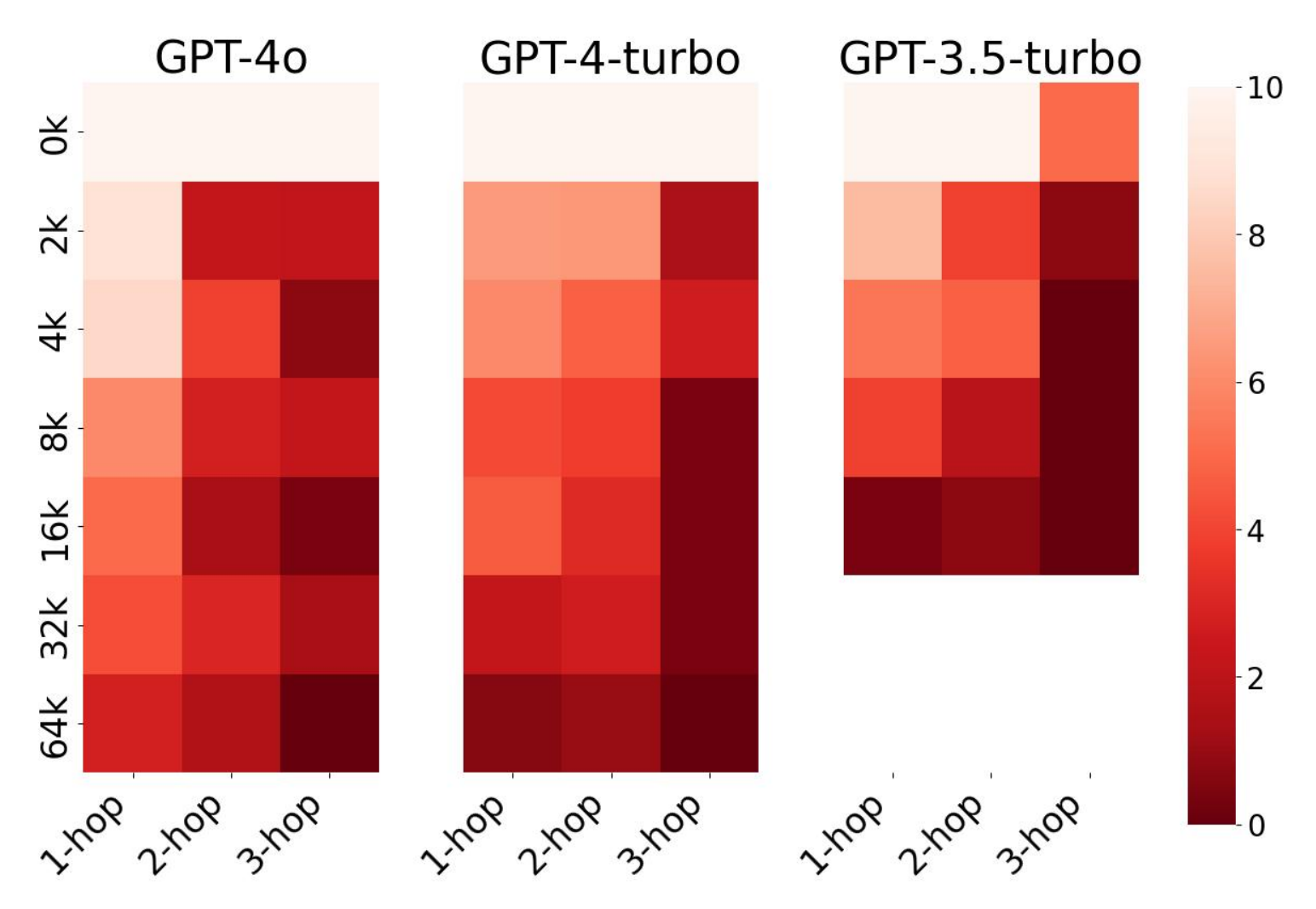}
                    \label{fig:multi}
                }
                \hfill
                \subfloat[Single-Needle (Combined at Depth 40\% or 60\%)]{
                    \includegraphics[width=0.48\linewidth]{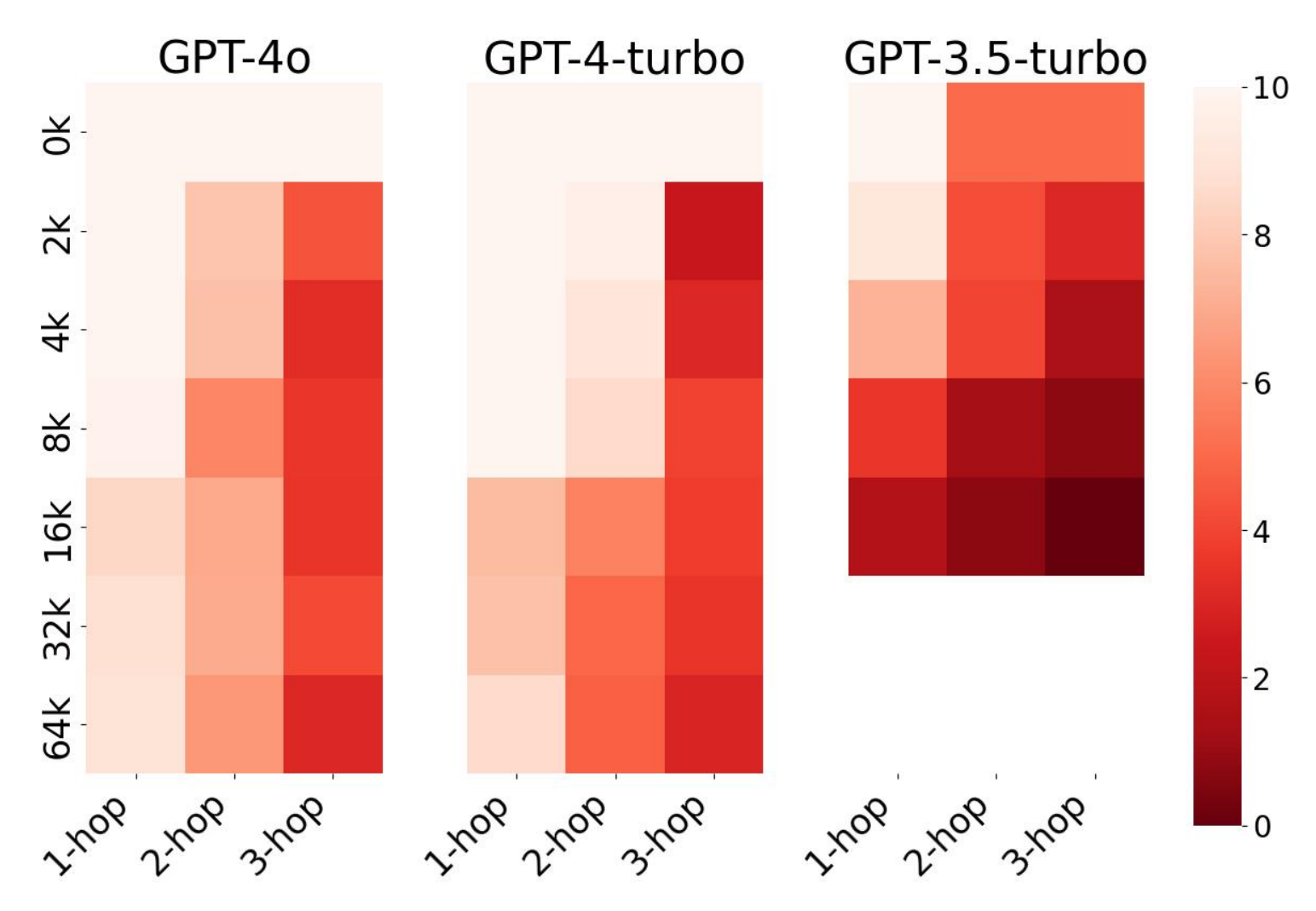}
                    \label{fig:single_avg}
                }
            \end{minipage}
        };
    \end{tikzpicture}
    \caption{Reasoning-in-a-haystack Comparison based on the Mebot's Real Data w.r.t. Different Context Lengths and Hop Counts. 
    The multi-needle setting distributes different needles uniformly in the haystack, and the single-needle setting merges all the needles together and injects them at either depth 40\% or 60\%. The scores are averaged across different runs.
    \texttt{GPT-3.5-turbo} cannot be applied to too long context lengths.
    Detailed results are illustrated in Figure ~\ref{fig:experiment_result_with_number}.} 
    \label{fig:experiment_result}
\end{figure}

As shown in Figure~\ref{fig:experiment_result}, the most recent LLMs from OpenAI, \texttt{GPT-4o} and \texttt{GPT-4-turbo} both show poor performance with long texts and multiple hops, supporting our aforementioned arguments on LLMs and AGI.
Checking the score trend over the number of hops and the context length,
it is obvious that the quality of responses is negatively correlated both of them, indicating that LLMs struggle with extended texts and multiple reasoning steps.
Also, the results confirm that the multi-needle setting is more challenging than the single-needle one, because combining all the needles together reduces the retrieval difficulty. 

Remarkably, \texttt{GPT-4o} and \texttt{GPT-4-turbo} perform similarly on this task.
According to livebench results (\url{https://livebench.ai/}), \texttt{GPT-4-turbo} outperforms \texttt{GPT-4o} in reasoning tasks, while \texttt{GPT-4o} excels in language tasks. 
Since our task combines these two aspects, similar results for both models are consistent with the literature.

\subsection{Remarks and Discussions}
The current reasoning ability of LLMs is insufficient.
Without a new paradigm that significantly improves reasoning ability, it is very unrealistic to rely entirely on LLMs to complete all necessary retrieval and reasoning in one step.

Drawing the connections with human learning and reasoning, the context of LLMs is like short-term (working) memory. 
Even with super long/unlimited effective long-context LLMs, they can only solve problems based on very long short-term memory -- every time, the LLMs work everything from the scratch.
Intuitively, this is less efficient and effective than saving and organizing the important conclusions from the history. 
Therefore, the most ideal approach here is to timely transform important conclusions into long-term memory for better future use. 
This points us to AI-native memory.
\section{AGI should be a System with AI-Native Memory}

AGI shall be a System like a computer, where LLMs are like Processors and the context of LLM is like RAM.
To complete this system, we must have (long-term) Memory as disk storage. 

\smallsection{RALM/RAG is an elementary version of Memory}
Retrieval-augmented LLMs (RALMs) that sift through numerous relevant contexts to answer a query~\citep{kovcisky2018narrativeqa,dasigi2021dataset,pang2022quality,trivedi2022musique} can be viewed as a special case here by defining the Memory as raw data only. 
While some people want to leverage RALMs for AGI, the main starting point of these methods were to solve the lack of domain knowledge in LLMs.
Therefore, these methods are designed to solve the problem that the long-context supported by LLM itself is not long enough. 
As discussed earlier, relying solely on the super long context of LLM itself cannot realize AGI. 
So RALM/RAG doesn't work either.

Memory is beyond the raw data, as it should be generated and organized, including many results that require reasoning from the raw data. 
In addition to downstream applications, Memory shall be able to be directly consumed by users. 

\noindent\textbf{What is AI-Native Memory?}
We believe the ultimate form of AI-Native Memory is a deep neural network model (thus \emph{AI-native}) that parameterizes and compresses all types of memory, even the ones cannot be described by natural languages.
In order to ensure the privacy of the Memory across different users who interacted with the same AGI agent, we argue that the best practice is to maintain one Memory model for each individual user. 
Therefore, we refer to this Memory model between the AGI agent and a particular user as the \textbf{Lifelong Personal Model (LPM)} of this user. 
The LPM records, organizes, indexes, and arranges every detail about the individual, ultimately providing interfaces for users to directly access memories and for downstream applications (such as personalized generation, recommendations, etc.) to utilize useful, complete contexts. 
In a sense, the LPM acts as an upgraded ``Retrieval-Augmented'' role. 
Its superiority lies in the transformation of original data through extensive ``reasoning'' (i.e., organizing, indexing, etc.), rather than merely recording. 
Note that the LPM will evolve as the user interacts with LPM, creating a \emph{Data Flywheel}.

We envision three levels of the implementations of LPM as follows, with increasing complexity.
\begin{itemize}[nosep,leftmargin=*]
    \item \textbf{L0: Raw Data}. This approach is similar to directly applying RALM/RAG to raw data, defining Memory as all raw data. 
    \item \textbf{L1: Natural-language Memory} refers to the memory that can be summarized as natural language forms, such as short bio of the user, a list of significant sentences or phrases, and preference tags.
    \item \textbf{L2: AI-Native Memory} refers to the memory that doesn't necessarily need to be described in natural language, learned and organized through model parameters. 
    Each LPM will be a neural network model.
\end{itemize}
From a technical perspective, the production, organization, consumption, and maintenance of the LPM need to be addressed.
The rest of this section will give a deep dive into L1 and L2.

\subsection{L1: Natural-language Memory}

In L1, the Memory will include a set of natural-language descriptions, such as keywords/tags, phrases, sentences, and even paragraphs. 
These are highly relevant to information extraction and knowledge discovery, including phrase mining~\citep{shang2018automated,gu2021ucphrase}, entity recognition~\citep{peng2023less}, relation extraction~\citep{hogan2022fine}, text summarization~\citep{widyassari2022review}, taxonomy construction~\citep{tao2018doc2cube,shang2020nettaxo}, etc. 
It will also cover different modalities as sources of the Memory, such as image, audio, video, and even sensor signals from wearable devices. 

The developers of the L1 LPM have to specify the schemes. 
For example, various useful Memory types can be defined, including but not limited to
\begin{itemize}[nosep,leftmargin=*]
    \item \emph{(Short) Bio}, a general description of the user, typically with a few sentences. 
    \item \emph{Topics} of interest to the user, which can be seen as a collection of tags (e.g., ``politics'', ``basketball'').
    \item \emph{Preferences} include a user's preferences for various things. The preference and topic are different because knowing a preference typically (implicitly) excludes the other side of the preference (e.g., detailed vs. concise expressions, cost-effective vs. luxury products, aisle vs. window seats). 
    \item \emph{Social Connections} include the user's social relationships, such as who and which organizations have been mentioned.
\end{itemize}

The Memory can be categorized by granularity too. Taking the topics as example, we can have the following examples from fine-grained to coarse-grained. 
\begin{itemize}[nosep,leftmargin=*]
    \item \emph{Summarized Sentences}: Each interaction with the user can be summarized into sentences. 
    Such summaries are just one level beyond the raw data in L0. 
    There can be redundancies, but they should not contradict each other.
    \item \emph{Fine-grained Tags}: Very precise tags that summarize Memory at a very detailed level. These tags are typically explicitly mentioned by the user.
    \item \emph{Coarse-grained Tags}: Starting from fine-grained tags, one can roll up the granularity to obtain more general tags. 
    For example, expanding from a player's name (e.g., Michael Jordan) to the sport league (e.g., NBA), and from the sport league to the sport itself (e.g., Basketball). 
    It is important to keep the granularity not too far from the original fine-grained tags, so the user would be still interested.
    \item \emph{Global}: Every user should have a high-level summary, similar to what the user would say during ice-breaking sessions. 
    This includes fun facts, personal hobbies, etc. 
\end{itemize}

The Memory is never only about (generalized) extractions. It requires more complex inference and reasoning. 
\begin{itemize}[nosep,leftmargin=*]
    \item Memory can include information that is inferred from a single conversation, for example, through summarizing and reflecting.
    \item Memory can be derived from cross-session interactions. 
    This is essentially a pattern mining -- deducing global information through user behavior from a few interactions. 
    This can be achieved through sampling and chaining by tags/sentences, and then run an LLM inference.
    For example, one can put all recent Memory about basketball and then ask an LLM to find a trend.
\end{itemize}
\subsection{L2: AI-Native Memory}

In L2, the Memory goes beyond the natural language forms and becomes a neural network model, and therefore, we name it as ``AI-Native''. 
This model aims to encode all the memories of the user.
The L2 LPM can be viewed as a personalized version of world models~\cite{matsuo2022deep}. 
It shall be able to predict the user behavior based on the user history.
To this extent, the L2 LPM can also make suggestions when the user is adding new inputs like an auto-completion. 
Note that L2 is not simply a parameterized version of L1. 
It shall generalize to more subtle patterns that cannot be defined by the system designers. 
It is an end-to-end solution without handcrafted schemes.
One can expect that ``prompting'' the L2 model can obtain the information that the developers can define in L1. 

\smallsection{Privacy and Security}
Our envisioned LPM separates the user history as all the LPMs are trained independently, so there is no concern that the LPM will leak the user's information to others.
The data and model security is another thing to pay attention to.

\smallsection{An LLM can be L2 LPM}
The memory encoding in L2 can be viewed as a compression of the raw data as lossless as possible.
From this compression perspective, choosing LLM as LPM and \emph{continuing the language modeling objective training} on the user history become a very intuitive solution. 
At the same time, finding the underlying patterns from the memory so the model will be able to generate novel reasoned memories/preferences is an important feature of L2. 
One can expect that an L2 model should be able to generate all the L1 memory.
Therefore, we can \emph{leverage the L1 results} as additional data for a \emph{supervised fine-tuning} of the L2 model. 
In summary, one can obtain an LPM via a combination of language modeling ``pre-training'' and instruction-following ``fine-tuning'' based on the user history. 
Remarkably, one shall be able to prompt the L2 LPM to uncover all the L1 information because the model will likely hit nearly a zero training error. 

\smallsection{Challenges and Potential Solutions}
There are several challenges and open problems require more research and thinkings. 
\begin{itemize}[nosep,leftmargin=*]
    \item \textbf{Training Efficiency}. 
    One intuitive but computationally complex method is for each user to fine-tune their own LLM.
    A possible implementation would involve learning how to generate Memory from raw data and how to produce the required Memory based on the current context within an end-to-end neural network model.
    A compromise method is to use LoRA~\cite{hu2021lora} to fine-tune a personal LLM for each user.
    Our initial experiments suggest that a LoRA 7B model is enough to capture the memory of a single user, as the training data size is several magnitudes smaller than the typical pre-training data size of LLMs. 

    \item \textbf{Serving Efficiency}. 
    As more L2 LPMs deployed for users, new infrastructure is needed for serving these models. This is more challenging than serving one single generic LLM for all the users.
    LPMs have been customized for different users.
    One advantage of using LoRA models is that different LPMs can still share common layers in the neural architecture. 
    We plan to develop a new serving framework that combines the computations in the common layers of different LoRA models, so the concurrent queries can be put into batches to increase the throughput and also reduce the serving cost. 
    Another direction to explore is to offload the L2 LPM serving to the user's edge device, e.g., a smart phone, after we quantize the model. 
    
    \item \textbf{Cold Start}, as a common problem in training deep neural networks, 
    is a straightforward challenge in L2. 
    We argue that L2 LPM should be only trained when the user has accumulated sufficient data.
    Otherwise, one can always roll back to L1 to offer some initial personalization experience.
    Another idea is to find some role-play methods~\citep{peng2024quantifying} to generate synthetic data to lower the entry bar of the L2 LPM for users.
    
    \item \textbf{Catastrophic Forgetting and Conflicts Resolving}. It is important to ensure that new memory is learned while preventing catastrophic forgetting of old memory.
    There are also cases where the newly added, correct memory should override the previous wrong information.
    There are already some pioneer researches along this line~\citep{wang2024memoryllm,wang2024augmenting,chang2024large}. 
\end{itemize}

\section{Our First Attempt to L2 LPM}

In this section, we introduce our first attempt to build an L2 LPM model via fine-tuning an LLM.
The most important thing is to synthesize high-quality, diverse training data to cover different aspects of LPM's capability, because the raw data from the user history of interacting with me.bot is not directly applicable.
We choose to leverage the most powerful LLMs, such as \texttt{GPT-4o}, for date generation.
Our experiences indicate that it is important to ensure both the prompts and answers in the synthesized training data do not follow trivial patterns (e.g., simple templates when asking questions); 
otherwise, the LLM (e.g., a 7B one) can quickly master all the data without learning any meaningful information. 
Also, chain-of-thought (COT) style answers are preferred, so the LPM can not only learn the user-specific knowledge, but also distill some abilities from \texttt{GPT-4o}.

\subsection{Me-following Capability}

Different from the instruction-following capability, LPM requires a ``me-following capability'' -- when there is no explicit mention, instruction, or information about ``me'' (i.e., the user), LPM needs to act just like it has all the user history and the right instruction in its context. 
This way, one shall be easily distinguish LPM's responses from those general LLMs.
To realize this, the key is to make the user feel the ownership of this model.

We first propose to introduce a special \texttt{<|ME|>} token, which is added at the beginning of all the user prompts to the LPM, including both training and testing prompts. 
During the fine-tuning, the LLM will learn a different generation style associated with this \texttt{<|ME|>} token. 
At the inference time, this \texttt{<|ME|>} token will lead us to user-specific generations.

Another factor to consider is that in LPM's use cases, the user will likely ask questions more frequently using the words like ``I'' and ``me''.
In contrast, LLMs' instruction-tuning data only has a very small portion of ``I'', for example, in OLMo~\citep{groeneveld2024olmo}, this portion is only about 10\%.
As a stopword, this is truly a small ratio. 
More importantly, we observe that most of the ``I'' occurrences are in quotation, task description, and translation cases.
Therefore, we propose to further rephrase the user prompts in the fine-tuning data to emphasize more on ``me'', so the user can prompt LPM more naturally.

\subsection{Data Generation and Augmentation}

Here, we introduce more details of our data generation for different scenarios and aspects.

\smallsection{Memory Retrieval (and Summary)}
One of the most popular usages of LPM is to retrieve the memory which the user noted in the me.bot.
Depending on what to retrieve, there are two typical scenarios.
\begin{itemize}[nosep,leftmargin=*]
    \item \textbf{Note Chunk Retrieval}. When the user mentions one or more specific keywords/keyphrases from the same/siebling chunk(s), the LPM shall be able to generate the exact or paraphrased chunk(s) as the answer.
    \item \textbf{Cross-note Retrieval}. It is reasonable to expect the LPM to generate a comprehensive answer according to all the chunks which include the same keyword/keyphrase.
    The answer is not simply a concatenation of those relevant chunks; instead, it should be a digested version. In this case, the COT-style answer would be more intuitive as it can first echo the original chunks and then make a summarization. 
\end{itemize}

\smallsection{Time-awareness}
Understanding dates and relative time periods is crucial in me.bot. 
An LPM should be able to retrieve the memory according to the date (e.g., ``What did I note on 2024-07-18?'') mentioned in the user prompt or a time period (e.g., ``yesterday'', ``last week'', and ``last month'').
Specifically, there are at least two types of queries that one can expect:
\begin{itemize}[nosep, leftmargin=*]
    \item \textbf{Date Retrieval}. Given a specific date, LPM is expected to retrieve (and summarize) the notes on that date. 
    \item \textbf{Period Summarization}. Given the current time and a language like ``yesterday'' and ``last week'', LPM is expected to retrieve and summarize the notes within the date range. The summarization here can also be detecting a trend. 
\end{itemize}

Existing LLMs are not good at understanding time-related concepts.
As shown in Table~\ref{tab:time_hard}, \texttt{GPT-4o} cannot even finish a date retrieval prompt preciously. 

To address this challenge, we propose to treat date and time as long keyphrases.
We will design particular chain-of-thought patterns to enforce the LPM understand the relative time period, and then make the time period retrieval similar to the keyphrase retrieval. 

\begin{table}[t]
    \centering
    \caption{A typical badcase using \texttt{GPT-4o} as an example.}
    \label{tab:time_hard}
    \begin{tabular}{l p{12cm}}
\toprule
    Prompt: & What happened on April 5, 2000? \\
    \midrule
    \texttt{GPT-4o}: & On April 5, 2000, the following significant events took place:

1. \textbf{Microsoft Antitrust Case}: U.S. District Judge Thomas Penfield Jackson ruled that Microsoft Corporation had violated U.S. antitrust laws by maintaining its monopoly position in the personal computer operating systems market through anticompetitive means. This ruling was a significant development in the case that had been brought against Microsoft by the U.S. Department of Justice and 20 state attorneys general. The judge's findings of fact laid the groundwork for subsequent legal and regulatory actions against Microsoft.

2. \textbf{Air Philippines Flight 541 Crash}: Air Philippines Flight 541 crashed near Davao City in the Philippines, resulting in the deaths of all 131 people on board. The Boeing 737-200 aircraft crashed into a coconut plantation while attempting to land in poor visibility conditions. This tragic accident was one of the deadliest aviation disasters in the Philippines' history.

These events marked notable moments in legal, business, and aviation history on April 5, 2000. \\
\midrule
    Explanations: & Both answers are wrong. There was no milestone of the Microsoft Antitrust Case on April 5, 2000, according to the facts we can find online (see Appendix~\ref{sec:microsoft} for the summarized details).
The nearest milestone was on April 3, 2000. 
Also, the Air Philippines Flight 541 crash occurred on April 19 instead of April 5. 

Moreover, the events that had a global impact on this date should include that Yoshiro Mori takes office as Japanese Prime Minister. 
This is not retrieved by \texttt{GPT-4o}. \\
\bottomrule
    \end{tabular}
\end{table}

\subsection{Training Framework}

In this very first attempt, we mainly focus on the supervised fine-tuning step.
Specifically, we follow the LoRA pipeline and fine-tunes LLM in a parameter-efficient way. 
Empirically, 5 epochs are enough for a well-trained LPM.

In the future, we plan to develop a second step, RLMF, reinforcement learning with ``me'' in the loop. The user is expected to provide (implicit) feedback to LPM via me.bot.
Another future direction lies in the incremental training of LPM.
For example, when the user is interacting with me.bot in a multi-turn conversation mode, the user definitely wants to update LPM in a live way, so in the next turn, LPM can provide responses conditional on the newly added message too. 

\subsection{Safety and Security}

Similar to the general instruction tuning of LLMs, our LPM model needs to go through a safety and security tuning, so the chosen LLM will not experience some catastrophic forgetting issue for its safety and security features.
For this part, we plan to reuse the relevant open-source datasets.

\section{Pilot Experiments}

In this section, we evaluate our L2 LPM, together with various RAG methods, based on one pilot user from me.bot. 
This pilot user has $538$ notes from 2024-04-13 to 2024-07-03. 
Some notes are transcribed from recorded audio of meetings and the original audio can be as long as one hour.

\subsection{Benchmark Design and Evaluation Metrics}

We have designed the benchmark data following four categories of reasoning-in-a-haystack questions.
These questions are categorized by the type of ``reasoning''.
Each category contains 15 different questions, so there are 60 questions in total.
\begin{itemize}[nosep, leftmargin=*]
    \item \textbf{Memory} refers to the questions that only require the model to retrieve the content related to certain entities/keywords or a time period,
    e.g., ``\emph{What did I do last week?}''. 
    The ``reasoning'' here is simple summarization.
    \item \textbf{Understand} refers to the questions that require the model to understand an abstract concept that never appear in the notes and infer based on relevant notes, e.g., ``\emph{What is the design principles of me.bot?}''
    \item \textbf{Predict} refers to the questions that require the model to predict some ``future'' behaviors for the user, e.g., ``\emph{What shall I do today?}''. 
    \item \textbf{Recommend} refers to the questions that require the model to make decisions according to the comprehensive understanding of the user, e.g., ``\emph{Given the following 10 books, which one would you recommend me to read first?}''
    Note that these book names may never appear in the user's notes (i.e., external information), so it requires very advanced reasoning capability of the model to make the decision. 
\end{itemize}
As one can see, these four categories largely follow an increasing order of difficulty levels, because they require more and more comprehensive and global understandings of the user. 
The only exceptions are those time-related Memory questions.
As mentioned before, LLMs in general cannot handle time-related information well. 

\smallsection{Metric}
For each question, we manually curated a reference answer.
For each category, we have its own judging criteria.
The answer and criteria have been double checked with the pilot user too to meet the expectation.
\texttt{GPT-4o} is introduced as a judge to automatically evaluate the answers generated by different methods. 
It will give a rating from 1 to 5 according to the human-specified criteria. 
In order to make sure this automated evaluation is reliable, we have also manually evaluated $240$ random samples from all the evaluated question-answer pairs.
The correlation between GPT and Human evaluation results is $0.9025$.
Therefore, we believe it is reasonable to trust the auomated evaluation results.

\subsection{Compared Methods}

We mainly compare our L2 LPM with RAG-based methods. 
\begin{itemize}[nosep,leftmargin=*]
    \item \textbf{RAG++} is an enhanced RAG framework by us.
    It is powered by elastic search for the initial retrieval.
    After the initial retrieval, we will rewrite the question with the help of \texttt{GPT-4o} conditioned on the top retrieved results. 
    Using the rewritten question, we further leverage embedding to refine the search and ranking.
    Moreover, the notes are all augmented by the summary generated from \texttt{GPT-4o}.
    The final generation is done by \texttt{GPT-4o}.
    \item \textbf{GraphRAG}~\cite{edge2024local} is arguably the most effective RAG framework in the literature.
    It first constructs a knowledge graph with entities and relations extracted from the documents, and then maps the questions to the most similar entities and relations to decide a scope of relevant contexts. 
    These contexts are then feed to an LLM to generate the answer to the questions. 
    GraphRAG has two variants, \textbf{GraphRAG-local} that focuses on the local contexts of the entities and relations, and \textbf{GraphRAG-global} that focuses on the global summaries of different clusters on the knowledge graph.
\end{itemize}

RAG can be done based on different datasets as follows.
\begin{itemize}[nosep,leftmargin=*]
    \item \textbf{Raw Notes} include all the original notes in me.bot. 
    \item \textbf{LPM Data} refers to the data generated for our L2 LPM training. The size of the LPM data is one magnitude larger than raw notes.
\end{itemize}

When applying GraphRAG to the LPM Data, considering the overwhelming cost, we extract the entities and relations using \texttt{GPT-4o-mini}.
To keep it consistent and comparable, we use \texttt{GPT-4o-mini} for GraphRAG on Raw Notes too.

We have also included long-context LLMs' results using \textbf{Gemini} 1.5 Pro and \textbf{GPT-4o}.
The prompts used in these long-context LLMs are the same as the prompts in GraphRAG.
Note that the number of the tokens in the raw data have already exceeded the context length in these LLMs, so we only keep the most recent data.
Because of such truncation, there are 5 (33\%) Memory questions that can never be answered correctly by these baselines.
For a fair comparison, we have excluded them when calculating the score for these two models. 

\subsection{Experiment Setup}

Since the pilot user mostly use Chinese, we choose \texttt{Qwen-2-7B-instruct}~\cite{qwen2}, one of the most effective LLMs in Chinese. 
In our LPM training, the LoRA rank is set to $64$, the fine-tuning epoch is set to $5$, and we adopt a cosine learning rate scheduler with the max learning rate $0.0001$. 
We set the decoding temperature as $0$ for more stable results.

\begin{table}[t]
    \centering
    \caption{Pilot Experiment Results. The ratings are from 0 to 5 and averaged among 15 questions in each category. From left to right, the four categories largely follow an increasing order of difficulty levels, because they require more and more comprehensive and global understandings of the user. The only exceptions are those time-related Memory questions. Please note that in \emph{long-context LLM baselines}, the raw data has been truncated so we have \emph{excluded 5 (33\%) Memory questions} that can never be answered correctly by these baselines; therefore, those two numbers with * are inflated, e.g., $3.1$* means $3.1 \times 10 / 15 = 2.07$. The \textbf{best} results are in bold and the \underline{second best} ones are underlined. }
    \label{tab:lpm_results}
\resizebox{\linewidth}{!}{
    \begin{tabular}{llcccccc}
    \toprule
    \textbf{Method} & \textbf{Data} &  \textbf{Memory} & \textbf{Understand} & \textbf{Predict} & \textbf{Recommend} & \textbf{Average}\\
    \cmidrule(lr){1-2} \cmidrule(lr){3-6} \cmidrule(lr){7-7}
    Long-context LLM-gemini (L0) & Raw Notes & 2.2 & 3.33 & 3.87 & 4.2 & 3.400\\
    Long-context LLM-gemini (L0) & Raw Notes (128k) & 2.2* & \underline{3.47} & 3.93 & \underline{4.07} & 3.417\\
    Long-context LLM-4o (L0) & Raw Notes (128k) & 3.1* & 3.4 & \underline{4.0} & 3.2 & \underline{3.425} \\
    \midrule
    RAG++ (L0) & Raw Notes & 1.73 & \underline{3.47} & 3.8 & 3.33 & 3.083\\
    GraphRAG-global (L0) & Raw Notes & 2.33 & 2.8 & 3.07 & 3.6 & 2.950\\
    GraphRAG-local (L0) & Raw Notes & 2.47 & 3.2 & 3.93 & 3.53 & 3.283\\
    \midrule
    GraphRAG-global (L1) & LPM Data & 2.73 & 3.13 & 3.87 & 3.53 &3.317\\
    GraphRAG-local (L1) & LPM Data & \underline{2.87} & 3.20 & 3.87 & 3.40 & 3.333\\
    \midrule
    Our LPM (L2) & LPM Data & \textbf{3.13} & \textbf{3.8} & \textbf{4.2} & \textbf{4.6} & \textbf{3.933}\\
    \bottomrule
    \end{tabular}
}
\end{table}

\subsection{Results}

Table~\ref{tab:lpm_results} presents our evaluation results.
Looking at the average evaluation score, our LPM achieves the best overall performance, and long-context LLMs secured the second best place.
It is worth mentioning that our LPM is only a 7B model while those long-context LLM baselines are several magnitudes larger. 
So our LPM shall be considered as a significantly more effective and efficient solution than long-context LLMs. 
It is a bit of surprise that the RAG methods cannot beat long-context LLMs, in terms of the average score.

Here are more detailed discussions on different evaluation dimensions.
\begin{itemize}[nosep, leftmargin=*]
    \item \textbf{Memory}.
        All the compared methods didn't achieve a high score in the Memory type of questions.
        This is because most Memory questions are time-relevant, which is intrinsically hard for LLMs.
        Our LPM significantly outperforms all the baselines on Memory.
        Also, RAG methods generally perform better than long-context LLMs.
        Note that RAG++ performs worse than long-context LLMs.
        This indicates that when the retrieved contexts in RAG were irrelevant, they may even hurt the generation.
    \item \textbf{Understand} and \textbf{Predict} questions are typically not associated with any specific piece of data in the user's history. 
        So these questions are more friendly to long-context LLMs, since there should be many relevant information pieces for them to pick up for reasonably good generation. 
        For similar reasons, the simple RAG++ method doesn't lose to the complicated GraphRAG on these two types of questions.
    \item \textbf{Recommend} questions are very challenging to RAG methods, 
        because these questions typically have no entities/relations that can be found in the user history. 
        This makes RAG methods fail to find useful contexts and end up with some random ones, hurting their later generation steps. 
        Our LPM, benefiting from our data generation and augmentation, generalizes much better than long-context LLMs on recommendation. 
\end{itemize}

In summary, RAG methods, by design and also verified in our experiments, are great at answering specific questions (i.e., local questions) with explicit mentions of entities and relations.
Long-context LLMs have advantages in offering a macro understanding of the user and answering high-level questions (i.e., global questions).
Our LPM is able to achieve the best performance on both local and global questions, compared with all these baselines.

\section{Conclusions and Outlooks}

In this paper, we highlight the limitations of LLMs in achieving AGI due to the impracticality of unlimited context length. 
We propose that AGI should function as a system where LLMs act as processors, their context as RAM, and memory as a disk. 
Efficient memory is crucial, and we suggest two solutions: (1) constructing a ``Memory Palace'' using Information Extraction/Generation techniques for structured storage, and (2) compressing memory into a neural network for efficient retrieval. 
These approaches can be combined to create a robust memory system for AGI.

In our vision, the Memory is strongly associated with the user, and at the same time, agnostic to the specific applications. 
We believe that in the future, an AGI agent will first interact with the AI-Native Memory and see if it can supply the necessary information. If not, it is the AI-Native Memory's job to interact with the real user to figure out more information.
Therefore, AI-Native Memory will be the core of all interactions and personalizations between users and AGI agents. 
Note that personalization here is not only traditional content recommendation, but a type of recommendation service that marks the beginning of the AI journey.

With an accurate and efficient AI-Native Memory, it will enable numerous applications, such as memory-augmented chat, recommendations, building situational memory, auto-completion for the user's input, and integrating personal models based on relationships in social networks.

In conclusion, there is a significant potential of AI-native memory as the transformative infrastructure for (proactive) engagement, personalization, distribution, and social in the AGI era, as well as the incurred privacy and security challenges with preliminary solutions.

\bibliographystyle{neurips_2023}
\bibliography{neurips_2023}


\appendix

\section*{Appendix}

\section{An Example of Multi-Needle Reasoning-in-a-haystack}
\label{sec:example-multi-needle}

\newtcolorbox{AIbox}[2][]{colback=white!90!gray, colframe=black!75!white, fonttitle=\bfseries, title=#2, #1}
\begin{figure}[H] 
\begin{AIbox}{Example}
    \textbf{Q\_id}: Multi-Q2 \\[0.5em]
    \textbf{Needles:}
    \begin{itemize}
        \item I don't particularly enjoy butter, skip it.
        \item I'd like to cook a medium-rare steak at home, how can I do that?
        \item How can I make bacon that is nicely crispy for breakfast?
        \item Try to avoid spicy food for me.
        \item How can I make fried chicken for Emmy? She loves it.
        \item Could you recommend a good recipe for a creamy mushroom risotto for me?
        \item Alice prefers butter rather than bacon, note this for me.
    \end{itemize}
    \vspace{0.5em}
    \textbf{Query:} Given my preferences for diet, please help me to extract and specify them into two categories, namely like and dislike, and give simple reasons. Please only extract my preferences rather than others. Try to think step by step. \\[0.5em]
    \textbf{True Answer:}
    \begin{itemize}
        \item \textbf{Likes:}
        \begin{itemize}
            \item Medium-rare steak
            \item Crispy bacon
            \item Creamy mushroom risotto
        \end{itemize}
        \item \textbf{Dislikes:}
        \begin{itemize}
            \item Spicy food
            \item Butter
        \end{itemize}
    \end{itemize}
    \vspace{0.5em}
    \textbf{Difficulty:} High \\[0.5em]
    \textbf{Hop:} 2-hop \\[0.5em]
    \textbf{Type:} Information Retrieval and Reasoning \\[0.5em]
    \textbf{Breakdown:}
    \begin{enumerate}
        \item Identify preferences in diet.
        \item From those preferences, extract the user's preference rather than others.
    \end{enumerate}
\end{AIbox}
\caption{Example Query-Needle Pair and its True Answer.}
\label{fig:user_preferences_q2}
\end{figure}

\section{Prompt Template in Reasoning-in-a-haystack Experiments}
\label{sec:prompt_template}

\begin{figure}[H]
\begin{AIbox}{Prompt Settings}
\tiny
\begin{verbatim}
[{
    "role": "system",
    "content": "You are a helpful AI bot that answers questions for a user. Keep your response short and direct"
},
{
    "role": "user",
    "content": {context}
},
{
    "role": "user",
    "content": f"{question} Don't give information outside the document or repeat your findings"
}]
\end{verbatim}
\end{AIbox}
\caption{System Prompt used in the Provider LLM.}
\label{fig:chat_prompt}
\end{figure}

\section{Evaluation Criteria in Reasoning-in-a-haystack Experiments}
\label{sec:eval_criteria}

\begin{figure}[H]
\begin{AIbox}{Criteria for evaluation}
    \textbf{Accuracy}: \\
    Score 0: The answer is completely unrelated or contradictory to the reference. \\
    Score 3: The answer has minor relevance with the reference but does not align with the reference. \\
    Score 5: The answer has moderate relevance but contains inaccuracies. \\
    Score 7: The answer aligns with the reference but has minor omissions. \\
    Score 10: The answer is completely accurate and aligns perfectly with the reference. \\
    Only respond with a numerical score
\end{AIbox}
\caption{Criteria used in the Evaluator LLM during Scoring.}
\label{fig:accuracy_scale}
\end{figure}

\section{Detailed Experiment Results of Reasoning-in-a-haystack}

\begin{figure}[t]
    \centering
    \begin{tikzpicture}
        \node[rotate=90] at (-5.8, 1.4) {\small Context length};

        \node at (0, 1) {
            \begin{minipage}{\textwidth}
                \centering
                \subfloat[Multi-Needle (Uniform Depth)]{
                    \includegraphics[width=0.8\linewidth]{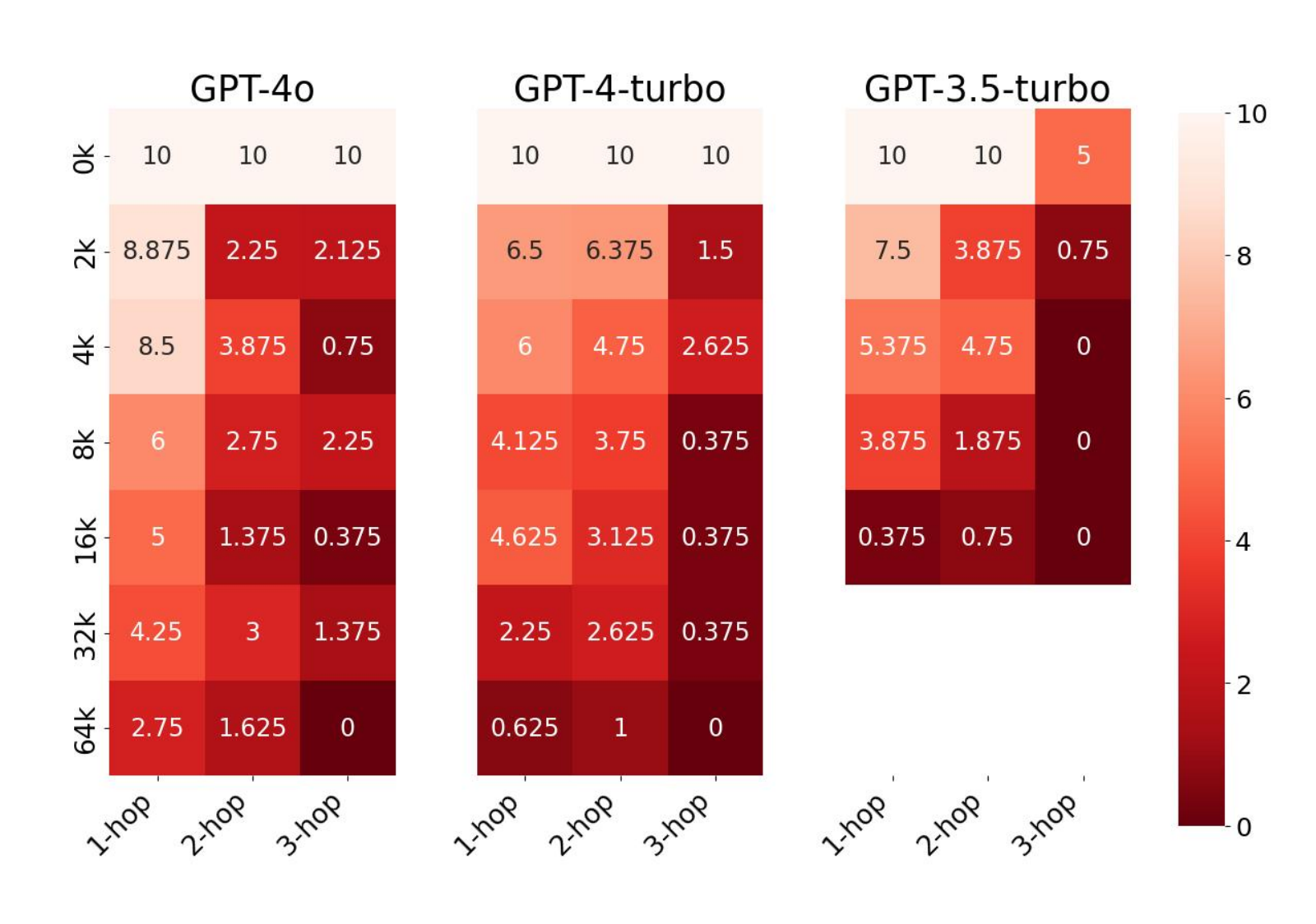}
                    \label{fig:multi_appendix}
                }
            \end{minipage}
        };

        \node[rotate=90] at (-5.8, -7.1) {\small Context length};
        \node at (0, -7.5) {
            \begin{minipage}{\textwidth}
                \centering
                \subfloat[Single-Needle (Combined at Depth 40\% or 60\%)]{
                    \includegraphics[width=0.8\linewidth]{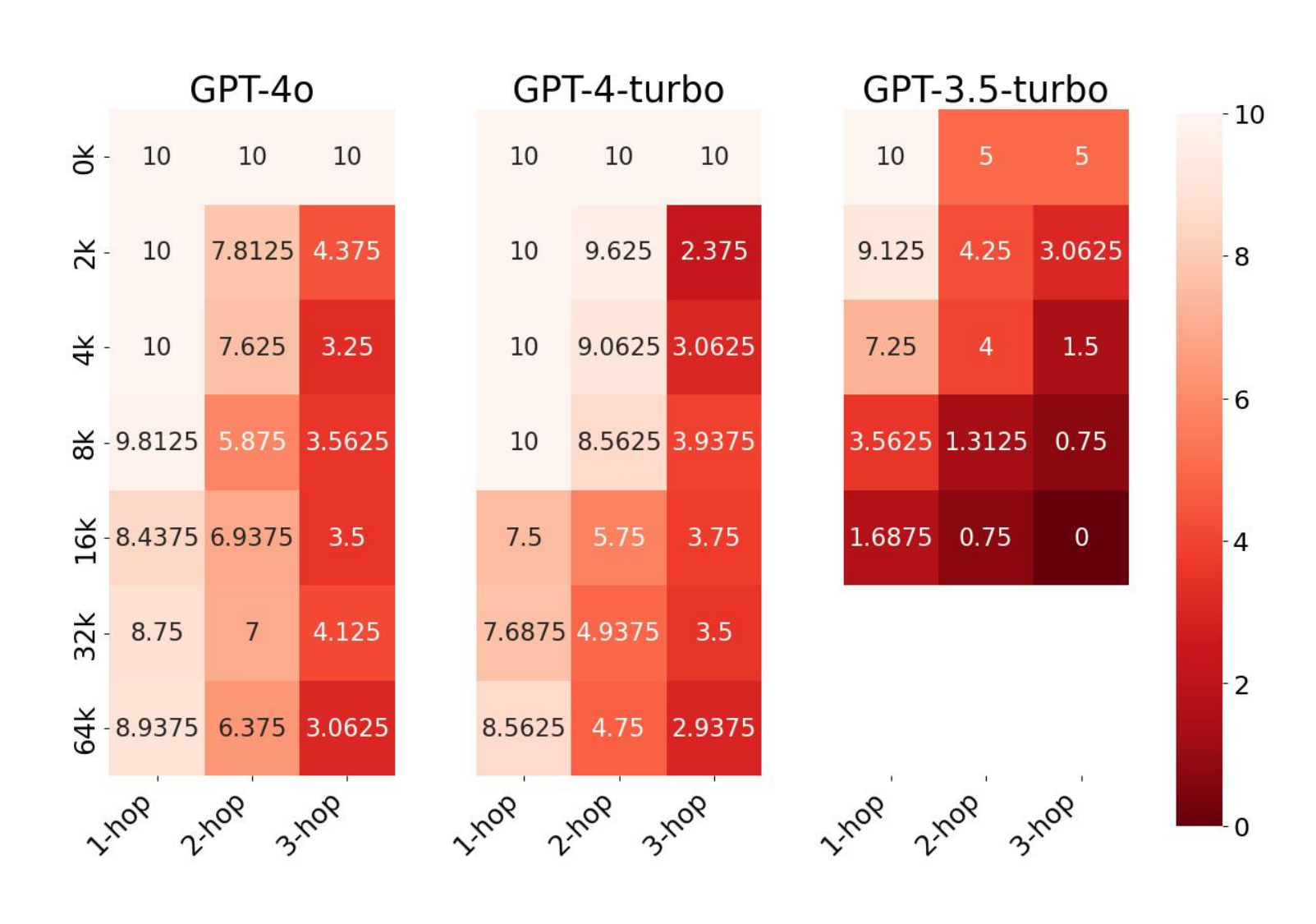}
                    \label{fig:single_avg_appendix}
                }
            \end{minipage}
        };
    \end{tikzpicture}
    \caption{A detailed version of Figure~\ref{fig:experiment_result} with specific numbers.
    }
    \label{fig:experiment_result_with_number}
\end{figure}

Figure~\ref{fig:experiment_result_with_number} shows the heatmaps with the specific numbers.

\section{The Antitrust Case against Microsoft}
\label{sec:microsoft}

The timeline of the key milestones in the antitrust case against Microsoft, presided over by Judge Thomas Penfield Jackson:
\begin{enumerate}
    \item \textbf{May 18, 1998}: The U.S. Department of Justice and 20 states filed an antitrust lawsuit against Microsoft, accusing the company of using its monopoly power in the operating systems market to eliminate competition, particularly in the web browser market.
    \item \textbf{October 19, 1998}: The trial officially began. Microsoft's CEO, Bill Gates, provided testimony via a pre-recorded video, explaining Microsoft's business practices and market strategies.
    \item \textbf{November 5, 1999}: Judge Thomas Penfield Jackson issued a preliminary ruling, finding that Microsoft held a monopoly in the operating systems market and had used this power to harm competition.
    \item \textbf{April 3, 2000}: Judge Jackson released detailed findings of fact, confirming that Microsoft had violated antitrust laws. The ruling noted that Microsoft employed exclusive contracts and other anti-competitive practices to maintain and extend its monopoly in the operating systems and web browser markets.
    \item \textbf{June 7, 2000}: Judge Jackson ordered the breakup of Microsoft into two separate entities—one responsible for the Windows operating system and the other for other software products. This ruling aimed to address Microsoft's monopolistic behavior.
    \item \textbf{June 28, 2001}: The U.S. Court of Appeals overturned the breakup order but upheld the finding that Microsoft had violated antitrust laws. The court remanded the case to a lower court for reconsideration of appropriate remedies.
    \item \textbf{November 2, 2001}: The U.S. Department of Justice and Microsoft reached a settlement agreement. Under the agreement, Microsoft had to implement a series of measures to ensure market competition but avoided being broken up.
    \item \textbf{November 1, 2002}: The federal district court approved the settlement agreement, bringing the long-running legal battle to an end.
\end{enumerate}
These milestones mark the key developments and resolution of the antitrust case against Microsoft.

\end{document}